\documentclass[pmlr]{jmlr}

\usepackage{longtable}

\usepackage{booktabs}
\usepackage[load-configurations=version-1]{siunitx} 

\theorembodyfont{\upshape}
\theoremheaderfont{\scshape}
\theorempostheader{:}
\theoremsep{\newline}

\jmlrvolume{302}
\jmlryear{2025}
\jmlrworkshop{Deep Learning Indaba 2025}

\title[Benchmarking Automatic Speech Recognition Models for African Languages]{Benchmarking Automatic Speech Recognition Models for African Languages}

\author{%
\Name{Alvin Nahabwe} \Email{alvin.n256@gmail.com}\\
\addr Makerere University Centre for Artificial Intelligence, Kampala, Uganda.
\AND
\Name{Sulaiman Kagumire} \Email{sulaiman.kagumire@gmail.com}\\
\addr Marconi Lab, Makerere University, Kampala, Uganda.
\AND
\Name{Denis Musinguzi} \Email{denismusinguzi2511@gmail.com}\\
\addr Marconi Lab, Makerere University, Kampala, Uganda.
\AND
\Name{Bruno Beijuka} \Email{brunobeijuka@gmail.com}\\
\addr Marconi Lab, Makerere University, Kampala, Uganda.
\AND
\Name{Jonah Mubuuke Kyagaba} \Email{kyagabajonah@gmail.com}\\
\addr Makerere University Centre for Artificial Intelligence, Kampala, Uganda.
\AND
\Name{Peter Nabende} \Email{peter.nabende@mak.ac.ug}\\
\addr Department of Information Systems, Makerere University, Kampala, Uganda.
\AND
\Name{Andrew Katumba} \Email{andrew.katumba@mak.ac.ug}\\
\addr Department of Electrical Engineering, Makerere University, Kampala, Uganda.
\AND
\Name{Joyce Nakatumba-Nabende} \Email{joyce.nabende@mak.ac.ug}\\
\addr Department of Computer Science, Makerere University, Kampala, Uganda.
\addr Makerere University Centre for Artificial Intelligence, Kampala, Uganda.
}

\begin{document}

\maketitle

\begin{abstract}
Automatic speech recognition (ASR) for African languages remains constrained by limited labeled data and the lack of systematic guidance on model selection, data scaling, and decoding strategies. Large pre-trained systems such as Whisper, XLS-R, MMS, and W2v-BERT have expanded access to ASR technology, but their comparative behavior in African low-resource contexts has not been studied in a unified and systematic way. In this work, we benchmark four state-of-the-art ASR models across 13 African languages, fine-tuning them on progressively larger subsets of transcribed data ranging from 1 to 400 hours. Beyond reporting error rates, we provide new insights into why models behave differently under varying conditions. We show that MMS and W2v-BERT are more data efficient in very low-resource regimes, XLS-R scales more effectively as additional data becomes available, and Whisper demonstrates advantages in mid-resource conditions. We also analyze where external language model decoding yields improvements and identify cases where it plateaus or introduces additional errors, depending on the alignment between acoustic and text resources. By highlighting the interaction between pre-training coverage, model architecture, dataset domain, and resource availability, this study offers practical and insights into the design of ASR systems for underrepresented languages.
\end{abstract}

\section{Introduction}
\label{sec:intro}
Automatic speech recognition (ASR) has advanced rapidly with the development of self-supervised learning and multilingual pre-training. Models such as wav2vec 2.0 \cite{baevski2020wav2vec}, XLS-R \cite{babu2021xls}, Whisper \cite{radford2023robust}, and MMS \cite{pratap2024scaling} have made it possible to build ASR systems for languages with limited labeled data. Despite these advances, African languages remain underrepresented in both training resources and systematic evaluation. With more than 2,000 languages spoken across the continent and many lacking transcribed speech data, building accurate ASR systems remains a major challenge. Beyond the scarcity of data, there is little comparative evidence on how different pre-trained models perform under varying amounts of training data and under different decoding strategies, making it difficult to guide practical choices in low-resource contexts.

Recent work has made progress in the development of ASR systems for African languages, but much of it remains narrow in scope. Many studies investigate a single model trained on fixed datasets, which provides only limited insight into cross-model comparisons. Comparative evaluations of hybrid and self-supervised approaches have examined up to 15 languages \cite{Ritchie2022LargeVS}, and some efforts have developed trilingual systems for languages such as Kinyarwanda, Swahili and Luganda \cite{elamin2023multilingual}. Other studies have benchmarked Whisper, MMS, and XLS-R on IsiXhosa in both zero-shot and fine-tuned settings \cite{blocker2025benchmarking}. Broader initiatives have introduced multilingual datasets and baseline results for 15 African languages using Whisper, XLS-R, and wav2vec2, supporting reproducibility and unified benchmarks \cite{abdou_mohamed_multilingual_2024}. New corpora for Luganda and Swahili have also been released with baseline results based on pre-trained models \cite{joycebuilding}. Although these contributions represent valuable progress, most focus primarily on error rates. There is still little systematic analysis across model types, data scales, and decoding strategies, particularly in low-resource conditions where design decisions can strongly influence performance.

This study advances the field by moving beyond error reporting to analyze why models behave differently in African languages. We examine how architectural design, pre-training coverage, and dataset characteristics interact to shape outcomes. Our analysis highlights where models are most data efficient, how scaling effects differ across languages, and under what conditions external language models improve or degrade performance. By linking model behavior to linguistic complexity and domain-specific features, we provide insights that extend beyond benchmarking and offer practical guidance for system development in underrepresented languages.

To address these gaps, we benchmark four pre-trained ASR models: Whisper, MMS, XLS-R, and W2v-BERT in 13 African languages. Each model is fine-tuned on progressively larger subsets of transcribed speech, ranging from 1 hour to 400 hours depending on available data. For XLS-R and W2v-BERT, we also evaluated the use of external n-gram language model decoding, while Whisper and MMS are evaluated using their standard decoding methods. In addition to reporting word error rates, we analyze model behavior under different data conditions, identifying where each model has advantages and where language model decoding is most beneficial. This provides a cross-model, cross-language, and cross-scale evaluation of ASR systems for African languages, offering practical guidance for model selection, data planning, and decoding strategies in low-resource contexts.

\section{Related Work}
\label{sec:related_work}
Recent advances in transfer learning have allowed the adaptation of pre-trained ASR models to low-resource languages, including those spoken in Africa~\cite{KHEDDAR2023110851}. Studies have demonstrated that self-supervised models like wav2vec 2.0 and XLS-R can significantly reduce word error rates (WER) when fine-tuned on small amounts of labeled data. However, most evaluations focus on a single model or language, limiting broader conclusions about performance trade-offs across different settings.

Several works have explored the role of training the size of the dataset in ASR for low-resource languages. For example, \cite{Mainzinger2024} found that MMS models outperform XLS-R for Mvskoke as more training data become available, while \cite{Meelen2024} report that transfer learning provides the greatest benefits in extremely low resource conditions, with diminishing returns as the data grows. However, these findings are based on isolated language settings and do not compare multiple model families side by side.

Language model decoding has also been shown to improve ASR performance in low-resource contexts. Although XLS-R yields modest gains from n-gram LM~\cite{Jimerson2023}, larger improvements have been reported in agglutinative languages~\cite{Orken2020}. However, the benefits of LM decoding appear to decrease as the quality of the acoustic model improves~\cite{Mainzinger2024}, and increasing the size of the LM does not always improve performance~\cite{liu-etal-2023-studying}. These mixed findings highlight the importance of evaluating the effects of LM under varying training conditions.

In addition, data quality and domain relevance have been shown to influence model performance more than quantity alone. For example, \cite{Mak2024} found that in-domain speech data consistently outperform larger but unrelated datasets. Other work has shown that synthetic augmentation can help mitigate vocabulary limitations and boost performance in environments with extremely low resources~\cite{Jimerson2018}.

Although these studies offer important information, most are limited in scope and evaluate a single model, language, or training condition. This work expands on existing literature by benchmarking four pre-trained ASR models in 13 African languages, multiple training scales, and both with and without language model decoding. The aim is to provide more comprehensive evidence to inform model and decoding choices in practical low-resource scenarios.

\section{Data}
\label{sec:data}

\subsection{Speech Data}
We use speech datasets covering 13 African languages drawn from a range of publicly available corpora. These datasets represent multiple speaking styles, including read, conversational, and descriptive speech. Table~\ref{tab:data_details} summarizes the corpora and associated speech styles. For Ewe, we obtained access to the Waxal dataset, which is currently in the process of being open-sourced by its creators.

\begin{table}[htbp]
\floatconts
  {tab:data_details}%
  {\caption{Details of the datasets and languages used, showing speech styles.}}%
  {\begin{tabular}{lll}
    \toprule
    \bfseries Dataset & \bfseries Language(s) & \bfseries Speech Style \\
    \midrule
    Google FLEURS & Wolof, Swahili, Luganda, Lingala & Read \\
    Common Voice & Swahili, Luganda, Kinyarwanda & Read \\
    NCHLT corpus & Zulu, Xhosa, Afrikaans & Read \\
    AfriVoice & Lingala, Shona & Descriptive \\
    ALFFA & Swahili, Wolof & Read \\
    Kallaama & Wolof & Read \\
    Jeli-ASR & Bambara & Spontaneous \\
    Asheshi & Akan & Conversational \\
    LRSC & Lingala & Read \\
    BIG-C & Bemba & Conversational \\
    Oza75 & Bambara & Read \\
    AMMI & Swahili & Read \\
    \bottomrule
  \end{tabular}}
\end{table}

\subsection{Text Data for Language Models}
To support language decoding for XLS-R and W2v-BERT, we compiled text corpora for 12 of the 13 target languages from publicly available sources covering domains such as news, health, education, agriculture, and religious texts. These corpora were used to train 5-gram language models with modified Kneser–Ney smoothing \cite{zhang-chiang-2014-kneser}. We chose a 5-gram model to capture wider context while remaining computationally efficient, and smoothing ensured better handling of rare and unseen word sequences. Kinyarwanda was excluded from both fine-tuning and LM training, since most of its available speech data was already included in the pre-training corpora of XLS-R and W2v-BERT. Table~\ref{tab:lm_stats} summarizes the word counts for each language.

\begin{table}[htbp]
\floatconts
  {tab:lm_stats}%
  {\caption{Word statistics of text corpora used for LM training.}}%
  {\begin{tabular}{ll}
    \toprule
    \bfseries Language & \bfseries Total Words \\
    \midrule
    Afrikaans & 61,863,750 \\
    Akan & 15,776,227 \\
    Bambara & 8,874,946 \\
    Bemba & 1,652,462 \\
    Ewe & 5,434,954 \\
    Lingala & 9,025,326 \\
    Luganda & 9,168,243 \\
    Shona & 5,366,168 \\
    Swahili & 22,247,952 \\
    Wolof & 53,000,509 \\
    Xhosa & 4,456,247 \\
    Zulu & 5,221,066 \\
    \bottomrule
  \end{tabular}}
\end{table}

\section{Models}
\label{sec:models}

We evaluated four state-of-the-art pre-trained ASR models: XLS-R, Whisper, W2v-BERT, and MMS. All four models are self-supervised or semi-supervised architectures trained on large-scale multilingual corpora. For each model family, we selected the smaller checkpoints (Whisper-small, XLS-R 300M, W2v-BERT 2.0 600M, and MMS-1B). This choice was primarily motivated by computational feasibility. For Kinyarwanda, we report results only for Whisper, MMS, and wav2vec2-large. We excluded XLS-R and W2v-BERT because the Kinyarwanda Common Voice dataset was included in their pre-training, which could lead to data leakage and inflated performance estimates.

\begin{enumerate}
\item \textbf{XLS-R}~\cite{babu2021xls}: An extension of wav2vec~2.0~\cite{baevski2020wav2vec}, XLS-R was pre-trained on 436k hours of speech covering 128 languages. Several of the African languages in our study were represented in its pre-training corpus. For example, Common Voice v6.1 contributed about 87 hours of Afrikaans, 91 hours of Swahili, 2k hours of Kinyarwanda, 72 hours of Luganda, and 24 hours of Shona. VoxLingua107 added 56 hours of Zulu and 3 hours of Luganda, while the BABEL corpus provided conversational telephone speech for multiple African languages, including Zulu. Given the extensive pre-training coverage of Kinyarwanda, particularly the 2,000 hours from Common Voice, we excluded it from XLS-R fine-tuning. We use the \texttt{facebook/wav2vec2-xls-r-300m} checkpoint.

\item \textbf{Whisper}~\cite{Radford2022}: An encoder-decoder transformer trained on 680k hours of labeled speech and speech translation in 97 languages. The pre-training data include transcribed audio scraped from the web, with only small amounts of African language coverage (e.g., Swahili contributed about 5.4 hours). Whisper supports multilingual transcription and zero-shot inference, making it robust to domain and noise variations. We use the \texttt{openai/whisper-small} checkpoint.

\item \textbf{W2v-BERT}~\cite{chung2021w2vbertcombiningcontrastivelearning}: W2v-BERT integrates contrastive predictive coding with masked language modeling to jointly learn acoustic and linguistic representations. It was pre-trained on 4.5M hours of unlabeled speech spanning 143 languages. Among our target languages, the pre-training included approximately 370 hours of Luganda, 364 hours of Swahili, 103 hours of Afrikaans, and 67 hours of Zulu. Since Kinyarwanda was also part of the pre-training corpus, we excluded it from fine-tuning to avoid leakage. We use the \texttt{facebook/w2v-bert-2.0} checkpoint.

\item \textbf{MMS}~\cite{pratap2023scalingspeechtechnology1000}: The Massively Multilingual Speech project extended wav2vec~2.0 to over 1,700 languages using about 500k hours of speech, primarily from publicly available religious readings and multilingual corpora. Relevant African data in pre-training included 2,000 hours of Kinyarwanda, 407 hours of Luganda, and 179 hours of Swahili from Common Voice v9; 64 hours of Swahili, 90 hours of Lingala, 30 hours of Shona, and 108 hours of Afrikaans from VoxLingua107; and additional Bible recordings from the MMS-lab dataset for all of our languages except Bemba. Zulu was also covered in the BABEL corpus (about 1,000 hours of conversational telephone speech). We use the 1B parameter \texttt{facebook/mms-1b-all} checkpoint.

\item \textbf{Language Models}: For XLS-R and W2v-BERT, we incorporated external decoding using 5-gram KenLM models trained on our text corpora with modified Kneser–Ney smoothing~\cite{heafield2011kenlm}. This improved decoding by leveraging prior knowledge of word sequences, which is especially helpful under low-resource conditions.
\end{enumerate}

\section{Experimental Approach}
\label{sec:approach}

\subsection{Data Preprocessing}
\label{data-processing}
All transcripts were normalized by lowercasing text, removing punctuation (except apostrophes), and mapping diacritics to base characters unless native to the language. Audio was filtered to durations between 2 and 30 seconds and resampled to 16\,kHz.

For evaluation, we ensured that test sets were consistent across all training sizes for each language. Where official splits were available, we used them directly. For languages without official splits, whether from a single corpus or multiple datasets, we created 5-hour validation and test sets using stratified random sampling. The same test sets were used throughout all training conditions to enable fair comparison.

\subsection{Model Training}
\label{training-details}
We fine-tuned all models using BF16 precision with activation checkpointing to reduce memory usage. Optimization used AdamW with gradient norm clipping and a linear learning rate schedule that included a 10\% warm-up phase. Learning rates were tuned per language, typically between 1e-5 and 1e-3. Early stopping was applied based on validation loss, with patience of 10 epochs and a minimum improvement threshold of 0.001. Batch size was 64, with gradient accumulation to maintain efficiency. A dropout rate of 0.1 was applied to attention, projection, and intermediate layers to prevent overfitting.

For XLS-R, MMS, and W2v-BERT, we froze the feature extractor and added a linear classification head. Whisper, in contrast, was fine-tuned end-to-end. All models were initialized from publicly available checkpoints on Hugging Face to ensure reproducibility. To examine data efficiency, we trained models on incremental subsets of 1, 5, 10, 20, 50, 100, and 200 hours. Swahili, which had the largest amount of transcribed speech, additionally included a 400-hour subset. For languages with less data, we used all available hours and created subsets up to their limits.

We also evaluated the impact of decoding with external n-gram language models for XLS-R and W2v-BERT. Specifically, we trained 5-gram KenLM models using the compiled text corpora described in Section~\ref{sec:data}, applying the same normalization pipeline. Decoding was performed with beam search using the \texttt{pyctcdecode} library. Whisper and MMS were evaluated with their default greedy or beam search decoders, as they do not support external LM integration.

Training was conducted on NVIDIA A40, L4, RTX A6000, and A100 GPUs with 24–80 GB VRAM. Larger datasets, particularly those above 100 hours, were trained on A100-equipped machines to reduce training time. Training duration varied by model, language, and dataset size: XLS-R converged relatively quickly, W2v-BERT and Whisper required more time, and MMS exhibited intermediate training times. These resource setups allowed for scalable and consistent experimentation across diverse language-resource conditions.

\section{Results}
\label{sec:results}

In this section, we report WER/CER results on how model performance scales with increasing amounts of transcribed data and the effect of incorporating n-gram language model decoding.


\subsection{Data Scaling}
Table~\ref{werbydatasetsize} reports WER across the four models for different training set sizes and languages. In general, all models show substantial reductions in WER as more transcribed data becomes available, but the magnitude and timing of improvements differ by language and model family.

\setlength{\tabcolsep}{3pt}
\begin{table*}[htbp]
\floatconts
  {tab:werbydatasetsize}%
  {\caption{WER/CER of the four ASR models across different training data sizes for all datasets.}\label{werbydatasetsize}}%
  {\scriptsize
  \centering
  \hspace*{-1.2cm}
  \begin{tabular}{|l|l|l|l|l|l|l|l|l|l|}
    \hline
    \bfseries Language & \bfseries Model & \bfseries 1h & \bfseries 5h & \bfseries 10h & \bfseries 20h & \bfseries 50h & \bfseries 100h & \bfseries 200h & \bfseries 400h \\
    \hline

     &  XLS-R  &38.62/8 &16.47/3 &13.31/3 &5.24/1 &2.79/1 & -&- &-\\
    Afrikaans  &W2v-Bert  &22.70/3 &13.79/2 &10.94/2 &9.23/2 &3.23/1 &- &- &-\\
    &Whisper  &26.38/4 &17.94/3 &14.02/3 &12.35/6 &2.11/1 &- & -&-\\
    &MMS &22.23/3 &13.04/2 &9.65/2 &8.48/1 &3.69/1 &- & -&-\\
    \hline
    &XLS-R  & 100/100&55.5/17.8 &39.9/12.7 & 34.0/10.7&29.8/9.4&28.0/8.7 & - & -\\
    Akan &W2v-Bert  & 45.6/15.3& 35.2/11.5& 32.2/10.5& 30.2/9.7& 30.0/9.5& 30.3/9.7&- &-\\
    & Whisper  & 62.0/23.9 & 45.2/16.9& 40.9/14.9& 38.4/14.6& 41.4/21.9& 30.3/10.9& - & - \\
    & MMS & 98.8/85& 37.4/12.1& 35.0/11.2& 32.8/10.3& 31.7/9.9& 31.1/9.5& - & -\\
    \hline
    & XLS-R  &74.2/34.7 & 51.7/24.7 &32.8/16.4 & 20.0/8.9& -& -& -&-\\
    Bambara   &W2v-Bert  & 71.1/32.5 & 50.8/24.1&44.1/20.3& 36.8/16.3&- & -& -&-\\
    &  Whisper  & 78.5/40.9 &53.0/28.3 & 45.4/28.3& 22.9/14.1&- & -&- &-\\
    & MMS & 72.2/36.7& 54.9/26.8&38.3/18.1&19.0/8.9&- &- &- &-\\
    \hline
    &XLS-R  &100.00/100 &100.0/100 &64.74/19 &49.63/14 &44.07/13 &42.48/12 & -&-\\
    Bemba  &W2v-Bert  &61.86/16 &49.95/14 &45.86/13 &43.76/12 &38.32/11 &36.20/10 &- &-\\
    &Whisper  &66.20/19 &50.72/14 &47.43/14 &43.84/13 &39.84/40 &37.32/11 &- &-\\
    &MMS &53.57/11 &47.51/9 &44.28/9 &42.25/8 &39.11/8 &37.07/8 &- &-\\
    \hline
    & XLS-R  & 100/100& 55.1/15.7 &44.1/12 &36.7/10.8 & 32.8/9.6& 31.2/9.1 & - &- \\
    Ewe  &W2v-Bert  &47.9/14.1& 37.3/10.8 & 35.4/10.4 & 34.4/10 & 34.1/9.8& 32.2/9.4&- &-\\
    &   Whisper  & 59.1/19.6& 47.6/15.2&47.4/17.3 & 46.4/20.4 & 36.0/11.2&34.5/10.6 &- &-\\
    &  MMS & 100/100  &45.4/13.1 & 38.0/11.1& 37.0/11&35.7/10.2 & 34.1/9.9 &- &-\\
    \hline
    &  XLS-R  &93.35/26.94 &72.11/17.33 &59.95/13.52 &53.96/11.65 &49.67/10.54 &42.87/8.43 &40.51/7.78 &-\\
    Luganda &W2v-Bert  &55.69/11.63 &52.01/10.69 &49.09/9.86 & 47.25/9.29 & 44.26/8.57 & 41.77/7.89 &39.75/7.41 &-\\
    &   Whisper  &98.91/37.34 &54.06/13.04 &48.15/11.92 &40.31/9.68 &30.06/7.54 &24.93/6.42 &20.45/5.54 &-\\
    &  MMS &58.99/15.86 &53.11/12.98 &51.24/11.85 &49.91/10.93 &48.49/10.21 &47.11/9.95 &42.69/8.56&-\\
    \hline
    &  XLS-R  &77.02/28.44 &30.18/9.05 &23.25/7.78 &22.16/6.7 &19.85/9.38 &18.82/5.69&-&-\\
    Lingala  &W2v-Bert  &33.36/11.52 &29.92/11.5 &29.33/10.9 &28.78/10.97 &24.65/9.35 &24.19/9.59 &- &-\\
    &Whisper  &43.29/18.72 &40.84/18.99 &32.63/13.78 &29.61/12.07 &28.23/11.41 &25.25/10.21 &- &-\\
    &MMS &51.3/20.978&32.43/10.96&27.61/9.31 &26.79/9.13 &24.59/8.48 &23.96/8.32 &-&-\\
    \hline
    &  wav2vec2-large &100/100&79.00/30 &68.05/25 &65.00/25 &42.04/14 &39.00/13&38.00/12&- \\
    Kinyarwanda &Whisper &100.00/100&70.07/27 &58.05/23 &55.12/22 &49.00/20 &36.43/12 &32.37/10&-\\
    &   MMS  &42.46/15 &37.25/7 &34.46/7 &30.00/6&28.39/6 &23.59/5 &19.44/5&- \\
    \hline
    &  XLS-R  &100.00/99 &47.40/8 &53.07/10 &36.77/6 &30.71/5 &25.83/4 &- &-\\
    Shona & W2v-Bert  &34.73/6 &28.91/5 &27.23/5 &25.12/4 &23.54/4 &22.56/4 &- &-\\
    &   Whisper  &61.24/17 &42.05/11 &37.30/12 &32.88/9 &26.75/7 &30.92/7 &- &-\\
    &  MMS &100.00/96 &30.28/5 &28.20/5 &26.30/4 &24.60/4 &23.70/4 & -&-\\
    \hline
    &  XLS-R  &58.23/18.6 &33.11/14.91 &31.19/12.06 &27.31/10.31 &26.36/9.97 &23.61/7.82 &16.11/5.32 &11.36/3.72\\
    Swahili &W2v-Bert  &32.36/10.08 &29.57/9.11 &28.91/8.62 &26.92/7.94 &24.45/7.22 &24.21/7.19 &18.86/6.08 &18.50/6.06\\
    &  Whisper  &51.33/17.29 &44.15/15.59 &38.24/14.94 &33.59/13.24 &28.18/11.17 &27.25/11.22 &17.21/7.72 &16.88/7.12\\
    &  MMS &24.81/7.09 &23.97/6.79 &23.75/6.72 &23.17/6.54 &21.55/6.14 &21.42/6.21 &20.99/5.95 &20.88/5.91 \\
    \hline
    &  XLS-R  &54.70/10 &19.45/4 &11.42/2 &10.20/2 &8.53/1 & -& -&-\\
    Xhosa  &W2v-Bert  &27.83/5 &17.14/3 &11.04/2 &6.49/1 &7.13/1 &- & -&-\\
    &   Whisper  &33.56/7 &15.57/4 &12.90/6 &5.25/1 &5.87/1 &- & -&-\\
    &  MMS &27.54/4 &15.38/3 &12.21/3 &9.07/2 &9.76/2 & -&- &-\\
    \hline
    & XLS-R  &86.37/72.5  & 37.57/19.13&33.8/13 &20.15/9.2 &- &- &- &-\\
    Wolof & W2v-Bert&58.53/24.1 & 43.5/16.07& 35.07/14.57 & 26.7/9.2&- &- &- &-\\
    &   Whisper  &63.87/27.23 & 44.87/16.33& 39.33/19.77&27.3/16.8 & -&- &- &-\\
    &  MMS &56.03/21.93 & 42.73/16.83& 34.03/14.5 & 20.50/9.1& -& -&- &-\\
    \hline
    & XLS-R  &45.31/8 &24.67/4 &22.51/4 &12.84/2 &9.69/2 & -&- &-\\
    Zulu & W2v-Bert  &28.02/5 &19.58/3 &20.38/3 &16.51/3 &7.89/1& -& -&-\\
    &  Whisper  &33.56/8 &20.35/4 &20.62/4 &21.57/8 &8.20/1 & -& -&-\\
    & MMS &22.72/3 &19.58/3 &17.47/3 &17.27/3 &12.58/2 & -& -&-\\

    \hline
  \end{tabular}}
\end{table*}

\begin{enumerate}
\item \textbf{XLS-R}: XLS-R shows strong scaling behavior in the low- to mid-resource range, with particularly sharp WER reductions in some languages. For example, Afrikaans improves from 38.62\% to 2.79\% and Xhosa from 54.7\% to 8.53\% within the first 50 hours, while Lingala decreases from 77.02\% to 19.85\% over the same range. These results highlight that XLS-R can learn effectively from relatively modest amounts of data when moving beyond extremely low-resource conditions.

At higher data levels, the gains vary considerably by language. Swahili continues to improve steadily from 26.36\% at 50 hours to 11.36\% at 400 hours, while Luganda shows slower progress beyond 100 hours, dropping only slightly from 42.8\% to 40.51\% between 100 and 200 hours. This suggests that although early fine-tuning is generally effective, the point at which performance begins to plateau is language-dependent. Factors such as linguistic complexity, orthographic representation, and overlap with XLS-R’s pre-training corpus likely contribute to these differences.

Overall, XLS-R achieves highly competitive results in several languages with less than 100 hours of transcribed data, but the degree of benefit from additional data depends strongly on the language. These diverse trajectories are illustrated in Figure~\ref{fig:xlsr_loglog}.

\begin{figure}[t]
\floatconts
  {fig:xlsr_loglog}%
  {\caption{Log plot of XLS-R WER versus training hours across 12 African languages (Kinyarwanda excluded).}}%
  {\includegraphics[width=0.9\linewidth]{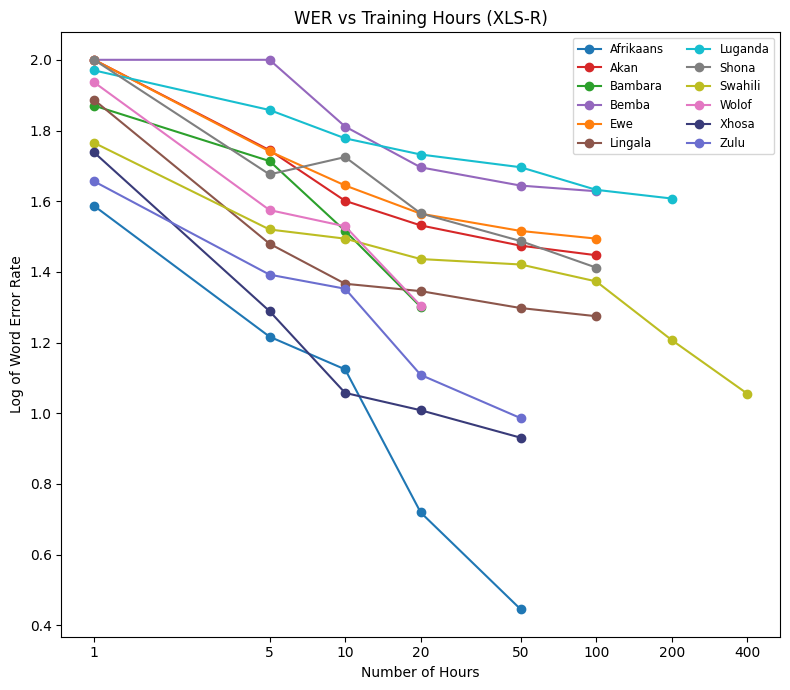}}
  \label{fig:xlsr_loglog}
\end{figure}

\item \textbf{W2v-BERT}: W2v-BERT demonstrates steady scaling across most languages, with strong early gains but slower improvements as more training data is added. For Afrikaans, WER decreases from 22.7\% at 1 hour to 3.23\% at 50 hours, showing high data efficiency in relatively simple settings. Xhosa follows a similar pattern, improving from 27.83\% to 7.13\% in the same range. Zulu also scales effectively, dropping from 28.02\% to 7.89\% by 50 hours. These languages show that W2v-BERT can reach sub-10\% WER with under 50 hours of fine-tuning.

For languages with larger datasets, improvements are more gradual. Swahili decreases from 32.36\% at 1 hour to 24.45\% at 50 hours, and then to 18.50\% at 400 hours. Luganda shows a similar trend, dropping from 55.69\% at 1 hour to 44.26\% at 50 hours, and then to 39.75\% at 200 hours. While both languages benefit from additional data, the rate of progress slows significantly beyond 50–100 hours, with diminishing returns despite substantial increases in training size.

Some languages plateau early, with little improvement even after 100 hours. Akan begins at 45.6\% and decreases to 30.0\% by 50 hours, but remains around 30\% thereafter. Ewe follows a nearly identical trajectory, dropping from 47.9\% to 34.1\% at 50 hours and then stagnating at 32.2\% by 100 hours. These patterns suggest that W2v-BERT struggles with certain linguistic properties or insufficient pre-training coverage for these languages.

Overall, W2v-BERT achieves reliable and consistent performance gains across African languages. It scales well in early training phases and reaches competitive WERs for Afrikaans, Xhosa, and Zulu, but offers diminishing returns beyond 50–100 hours in higher-resource cases. For Akan and Ewe, performance plateaus highlight limitations in fine-tuning effectiveness. These scaling behaviors are visualized in Figure~\ref{fig:w2vbert_loglog}.

\begin{figure}[h]
\centerline{\includegraphics[width=0.9\linewidth]{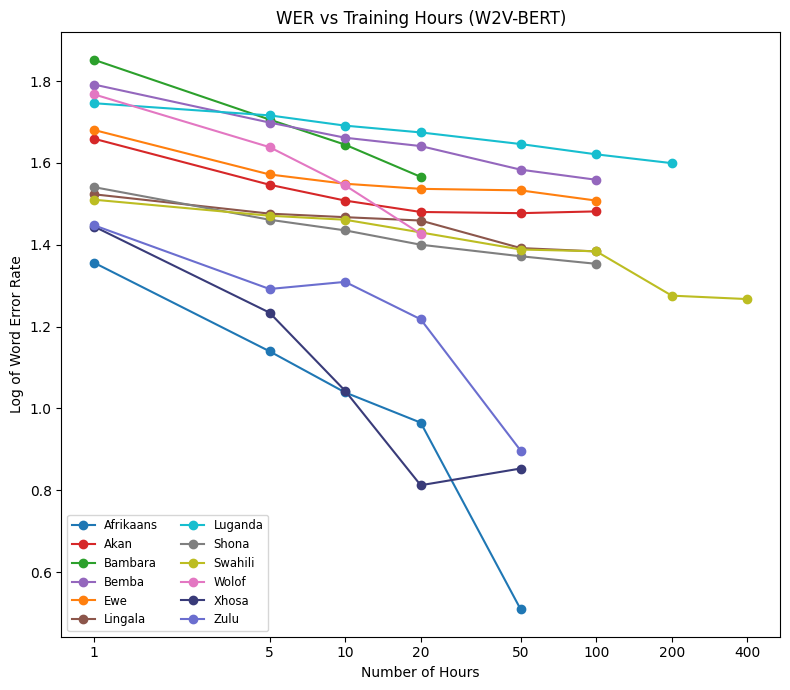}}
\caption{Log plot of W2v-Bert WER versus training hours across 12 African languages (Kinyarwanda excluded).}
\label{fig:w2vbert_loglog}
\end{figure}

\item \textbf{Whisper}: Whisper demonstrates rapid improvements in the early stages of fine-tuning, particularly for languages where it starts with very high error rates. For example, Luganda decreases from 98.91\% at 1 hour to 20.45\% at 200 hours, a relative reduction of almost 80\%. Afrikaans shows a sharper drop from 26.38\% to 2.11\% within 50 hours, and Xhosa follows a similar trajectory, falling from 33.56\% to 5.25\% in the same range. Zulu also improves from 33.56\% at 1 hour to 8.20\% at 50 hours. These results highlight Whisper’s strong data efficiency in certain languages during the first 50 hours of training.

At higher data scales, progress becomes more uneven. Swahili improves steadily from 51.33\% at 1 hour to 16.88\% at 400 hours, with diminishing gains after 100 hours. Shona declines from 61.24\% at 1 hour to 26.75\% at 50 hours, but fluctuates thereafter, ending at 30.92\% at 100 hours. Ewe plateaus at a high error rate (remaining above 30\% even at 100 hours), while Bemba decreases only modestly from 66.20\% to 37.32\% across 100 hours. Interestingly, Kinyarwanda, despite being included in Whisper’s pre-training, still reports high WERs, dropping from 100\% at 1 hour to 32.37\% at 200 hours, suggesting mismatches between the pre-training and fine-tuning data.

Overall, Whisper shows clear benefits from additional data in most languages, with the steepest gains concentrated below 50 hours. Beyond this range, some languages continue to improve steadily (e.g., Swahili), while others plateau or fluctuate, indicating that scaling effects are strongly language-dependent. These trajectories are shown in Figure~\ref{fig:whisper_loglog}.

\begin{figure}[h]
\centerline{\includegraphics[width=0.9\linewidth]{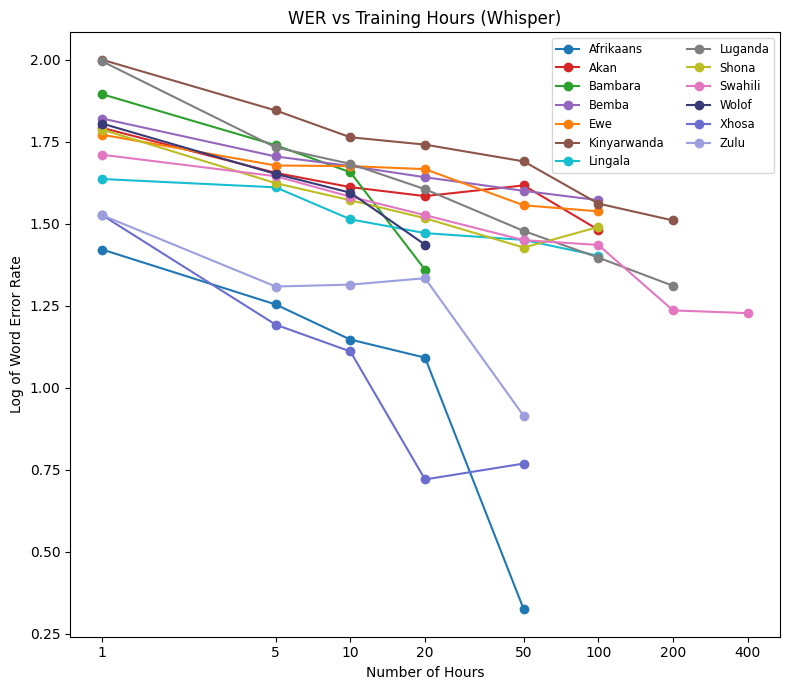}}
\caption{Log plot of Whisper WER versus training hours across 13 African languages.}
\label{fig:whisper_loglog}
\end{figure}

\item \textbf{MMS}: MMS shows strong scaling effects in the early phases of fine-tuning, particularly for low-resource languages. Bambara improves sharply from 72.2\% at 1 hour to 19.0\% at 20 hours, reflecting effective adaptation despite limited labeled data. Afrikaans follows a similar trajectory, dropping from 22.23\% to 3.69\% by 50 hours. These results suggest that MMS leverages its multilingual pre-training to achieve rapid error rate reductions with relatively small fine-tuning sets.

Other languages show more gradual progress. Bemba improves from 53.57\% at 1 hour to 37.07\% at 100 hours, while Luganda decreases from 58.99\% to 42.69\% across the same range. In both cases, performance gains taper off despite increasing training data, indicating a plateau effect. Similarly, Lingala improves from 51.3\% to 23.96\% between 1 and 100 hours, showing consistent but incremental progress.

For some languages, early improvements stall quickly. Ewe begins at 100\% and drops to 37.0\% by 20 hours, but improves only slightly afterward, ending at 34.1\% by 100 hours. Akan exhibits the same pattern, reducing from 98.8\% to 37.4\% at 5 hours but then stabilizing, finishing at 31.1\% at 100 hours. These plateaus likely reflect gaps in MMS pre-training coverage or the challenges of modeling tonal and morphologically complex languages.

Overall, MMS demonstrates clear benefits in lower-resource scenarios, with most gains achieved before 50 hours of training. Beyond this point, improvements slow or plateau, with outcomes varying significantly by language. Afrikaans and Bambara reach strong performance quickly, while Akan and Ewe remain stuck at relatively high error rates despite additional fine-tuning. These trajectories are shown in Figure~\ref{fig:mms_loglog}.

\begin{figure}[h!]
  \centering
  \includegraphics[width=0.9\linewidth]{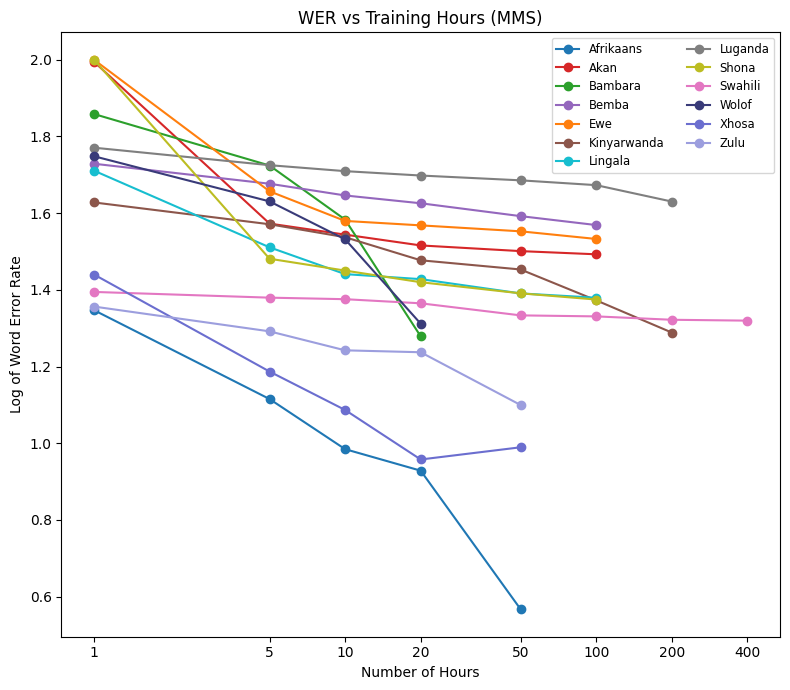}
  \caption{Log plot of MMS WER vs. training hours for MMS across 13 African languages.}
  \label{fig:mms_loglog}
\end{figure}

\end{enumerate}

\subsection{Language Modeling}

Table~\ref{tab:lm-table} compares XLS-R and W2v-BERT with and without external 5-gram language model decoding across training set sizes. Overall, LM decoding improves recognition performance in most African languages, particularly in low-resource settings below 50 hours. However, the size and consistency of these gains vary substantially across languages, and in some cases LM decoding leads to regressions.

\begin{enumerate}

\item \textbf{XLS-R}: LM decoding provides strong benefits for XLS-R in low- and mid-resource conditions. For example, Luganda improves from 93.35\% to 83.61\% at 1 hour and from 42.87\% to 20.68\% at 100 hours, with further gains to 17.43\% at 200 hours. Swahili shows similar improvements, dropping from 26.36\% to 20.28\% at 50 hours and from 11.36\% to 6.55\% at 400 hours. Afrikaans and Xhosa, which already perform well without LM, still benefit with small but consistent reductions, reaching 2.61\% and 7.78\% WER respectively at 50 hours. Substantial gains are also observed in Akan, where WER falls from 55.5\% to 40.7\% at 5 hours and from 34.0\% to 30.7\% at 20 hours. Wolof shows a similar pattern, dropping from 37.57\% to 32.6\% at 5 hours and from 20.15\% to 16.9\% at 20 hours. However, not all languages benefit. Bemba shows clear degradation, with WER rising from 49.63\% without LM to 60.72\% with LM at 20 hours, and remaining higher through 100 hours. Shona also experiences inconsistent gains, with WER dropping from 47.40\% to 39.95\% at 5 hours but increasing slightly from 30.71\% to 34.73\% at 50 hours. These results indicate that while LM decoding is particularly effective for XLS-R in the low- to mid-resource regime, its contribution is not uniform and can harm performance when the LM corpus is poorly aligned with the target language or domain (Figure~\ref{fig:xlsr-lm}).

\begin{figure*}[ht]
    \centering
    \includegraphics[width=\textwidth]{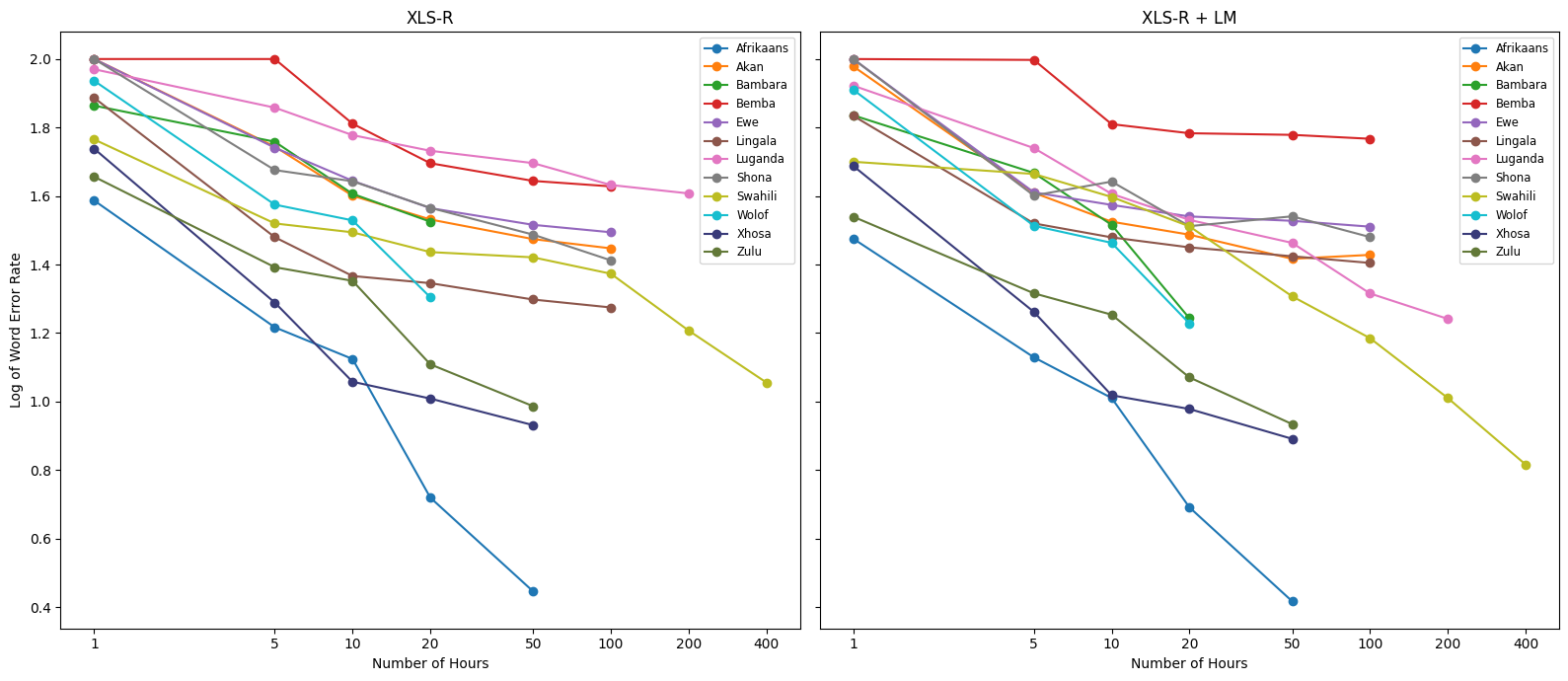}
    \caption{Comparison of XLS-R with and without language model decoding across varying training data sizes. WER values
are plotted on a log-log scale}
    \label{fig:xlsr-lm}
\end{figure*}

\begin{figure*}[ht]
    \centering
    \includegraphics[width=\textwidth]{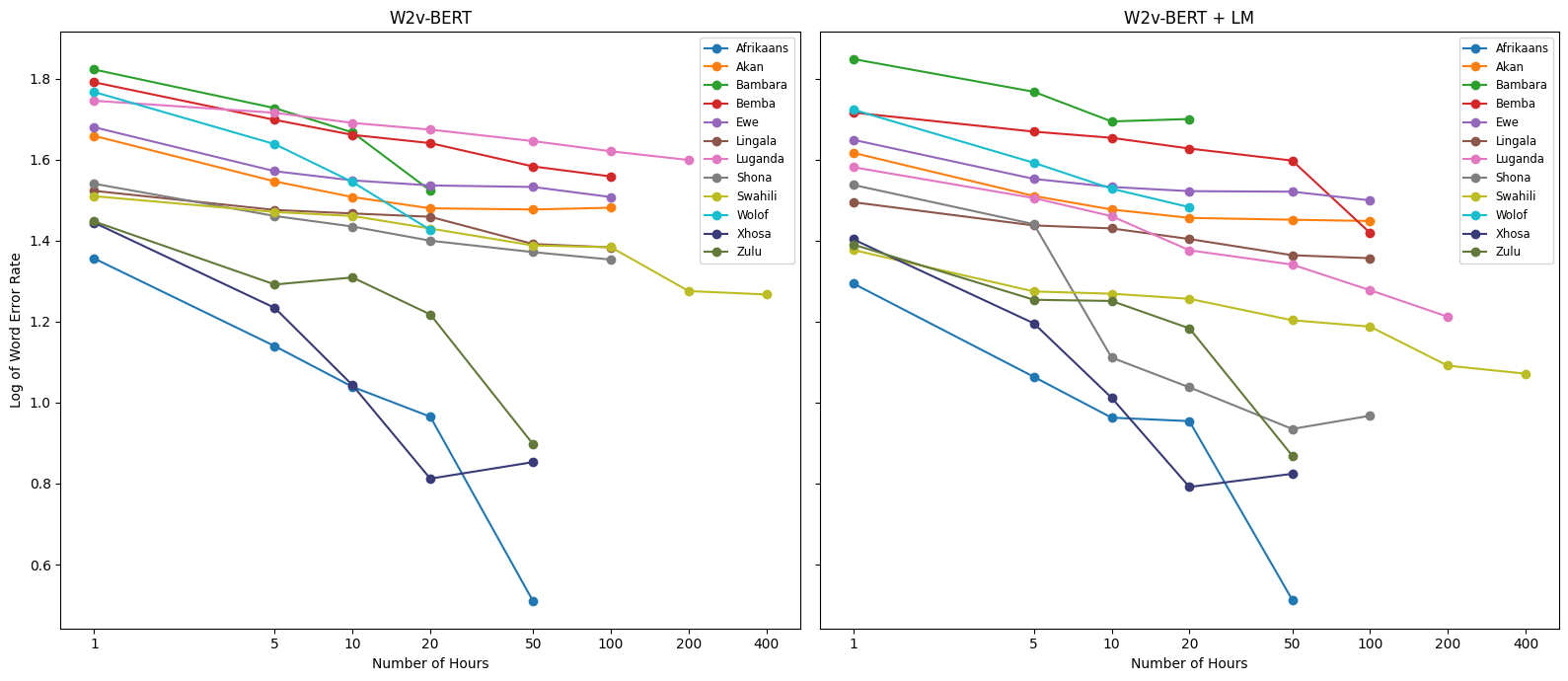}
    \caption{Comparison of W2v-BERT with and without language model decoding across varying training data sizes. WER values are plotted on a log-log scale.}
    \label{fig:w2vbert-lm}
\end{figure*}

\item \textbf{W2v-BERT}: W2v-BERT also benefits from LM decoding, especially in early and mid-resource conditions. In Luganda, WER decreases from 55.69\% to 38.16\% at 1 hour, from 44.26\% to 21.93\% at 50 hours, and from 41.77\% to 18.96\% at 100 hours. Swahili shows consistent reductions, dropping from 24.45\% to 15.98\% at 50 hours and from 18.50\% to 11.79\% at 400 hours. Afrikaans and Xhosa also achieve further improvements despite already low baseline error rates, reaching 3.26\% and 6.67\% WER, respectively, at 50 hours with LM. However, not all cases show improvements. In Bambara, WER initially improves slightly at 5 hours (50.8\% to 58.6\% with LM) but then worsens sharply at 20 hours, rising from 36.8\% to 50.2\%. Bemba, by contrast, shows strong late-stage gains, dropping from 36.20\% to 26.27\% at 100 hours. Akan and Ewe remain largely unaffected beyond 50 hours, with WERs stabilizing around 28–30\% regardless of LM integration. In general, W2v-BERT consistently benefits from LM decoding in languages with sufficient text–speech alignment, while other cases highlight that mismatched or limited LM resources can introduce noise and hinder recognition accuracy (Figure~\ref{fig:w2vbert-lm}).
 
\end{enumerate}

\setlength{\tabcolsep}{3pt}
\begin{table*}[h]
  \centering
  \caption{WER/CER of XLS-R and W2v-BERT with and without a language model (KenLM) across dataset sizes.\label{tab:lm-table}}
  \scriptsize
  \hspace*{-1.2cm}
  \begin{tabular}{|l|l|l|l|l|l|l|l|l|l|}
    \hline
    \textbf{Language} & \textbf{Model} & \textbf{1h} & \textbf{5h} & \textbf{10h} & \textbf{20h} & \textbf{50h} & \textbf{100h} & \textbf{200h} & \textbf{400h} \\
    \hline
    Afrikaans &XLS-R  &38.62/8 &16.47/3 &13.31/3 &5.24/1 &2.79/1 & -&- &-\\
              &XLS-R-LM  &29.88/5 &13.46/2 &10.24/2 &4.91/1 &2.61/0.5 &- &- &-\\
              &W2v-Bert  &22.70/3 &13.79/2 &10.94/2 &9.23/2 &3.23/1 &- &- &-\\
              &W2v-Bert-LM &19.69/3 &11.58/2 &9.18/2 &9.00/2 &3.26/0.5 &- &- &-\\
    \hline
    Akan &XLS-R  & 100/100&55.5/17.8 &39.9/12.7 & 34.0/10.7&29.8/9.4&28.0/8.7 & - & -\\
         &XLS-R-LM &95.2/89.5 &40.7/15.5 &33.5/11.6 &30.7/10.1 &26.1/8.8 &26.8/8.8 &- &-\\
         &WW2v-Bert  & 45.6/15.3& 35.2/11.5& 32.2/10.5& 30.2/9.7& 30.0/9.5& 30.3/9.7&- &-\\
         &W2v-Bert-LM &41.4/14.2 &32.4/10.9 &30.0/10 &28.6/9.4 &28.3/9.2 &28.1/9.4 &- &-\\
    \hline
    Bambara &XLS-R  &74.2/34.7 & 51.7/24.7 &32.8/16.4 & 20.0/8.9& -& -& -&-\\
            &XLS-R-LM &68.5/34 &46.5/23.6 &32.8/16.4 &17.5/8.8 &- &- &- &-\\
            &W2v-Bert  & 71.1/32.5 & 50.8/24.1&44.1/20.3& 36.8/16.3&- & -& -&-\\
            &W2v-Bert-LM &70.6/35.3 &58.6/30.1 &49.5/25.9 &50.2/26 &- &- &- &-\\
    \hline
    Bemba &XLS-R  &100.00/100 &100.0/100 &64.74/19 &49.63/14 &44.07/13 &42.48/12 & -&-\\
          &XLS-R-LM &100.00/100 &99.41/70 &64.54/23 &60.72/16 &60.66/15 &58.50/15 &- &-\\
          &W2v-Bert  &61.86/16 &49.95/14 &45.86/13 &43.76/12 &38.32/11 &36.20/10 &- &-\\
          &W2v-Bert-LM &52.07/17 &46.71/15 &45.10/13 &42.43/13 &39.62/13 &26.27/10 &- &-\\
    \hline
    Ewe &XLS-R  & 100/100& 55.1/15.7 &44.1/12 &36.7/10.8 & 32.8/9.6& 31.2/9.1 & - &- \\
        &XLS-R-LM &100/73.5 &40.8/13.3 &37.5/11.5 &34.7/11 &33.7/9.8 &32.4/9.6 &- &-\\
        &W2v-Bert  &47.9/14.1& 37.3/10.8 & 35.4/10.4 & 34.4/10 & 34.1/9.8& 32.2/9.4&- &-\\
        &W2v-Bert-LM &44.6/13.3 &35.7/10.5 &34.1/10.1 &33.3/9.8 &33.2/9.7 &31.6/9.4 &- &-\\
    \hline
    Lingala &XLS-R  &77.02/28.44 &30.18/9.05 &23.25/7.78 &22.16/6.7 &19.85/9.38 &18.82/5.69&-&-\\
            &XLS-R-LM &68.25/26.49 &33.12/13.98 &30.15/11.21 &28.19/10.54 &26.55/9.38 &25.39/9.25 &- &-\\
            &W2v-Bert  &33.36/11.52 &29.92/11.5 &29.33/10.9 &28.78/10.97 &24.65/9.35 &24.19/9.59 &- &-\\
            &W2v-Bert-LM &31.27/11.19 &27.40/9.91 &26.94/9.63 &25.36/9.15 &23.13/8.28 &22.74/8.52 &- &-\\
    \hline
    Luganda &XLS-R  &93.35/26.94 &72.11/17.33 &59.95/13.52 &53.96/11.65 &49.67/10.54 &42.87/8.43 &40.51/7.78 &-\\
            &XLS-R-LM &83.61/24.23 &54.96/13.65 &40.39/9.86 &33.91/8.02 &29.05/7.02 &20.68/4.8 &17.43/4.09 &-\\
            &W2v-Bert  &55.69/11.63 &52.01/10.69 &49.09/9.86 & 47.25/9.29 & 44.26/8.57 & 41.77/7.89 &39.75/7.41 &-\\
            &W2v-Bert-LM &38.16/8.32 &32.02/7.08 &28.88/6.29 &23.79/5.35 &21.93/4.89 &18.96/4.22 &16.30/3.72 &-\\
    \hline
    Shona &XLS-R  &100.00/99 &47.40/8 &53.07/10 &36.77/6 &30.71/5 &25.83/4 &- &-\\
          &XLS-R-LM &100.00/96 &39.95/8 &43.87/9 &32.51/9 &34.73/6 &30.21/5 &- &-\\
          &W2v-Bert  &34.73/6 &28.91/5 &27.23/5 &25.12/4 &23.54/4 &22.56/4 &- &-\\
          &W2v-Bert-LM &34.50/6 &27.60/5 &12.92/5 &10.91/3 &8.61/2 &9.28/2 &- &-\\
    \hline
    Swahili &XLS-R  &58.23/18.6 &33.11/14.91 &31.19/12.06 &27.31/10.31 &26.36/9.97 &23.61/7.82 &16.11/5.32 &11.36/3.72\\
            &XLS-R-LM &50.12/17.04 &46.18/15.25 &39.56/13.72 &32.47/11.51 &20.28/7.63 &15.31/5.97 &10.24/3.96 &6.55/2.94\\
            &W2v-Bert  &32.36/10.08 &29.57/9.11 &28.91/8.62 &26.92/7.94 &24.45/7.22 &24.21/7.19 &18.86/6.08 &18.50/6.06\\
            &W2v-Bert-LM &23.83/8.01 &18.82/7.93 &18.58/7.49 &18.05/6.97 &15.98/6.33 &15.41/6.29 &12.34/4.92 &11.79/4.92 \\
    \hline
    Wolof &XLS-R  &86.37/72.5  & 37.57/19.13&33.8/13 &20.15/9.2 &- &- &- &-\\
          &XLS-R-LM &81.33/71.3 &32.6/13.23 &29.06/12.53 &16.9/8.7 &- &- &- &-\\
          &W2v-Bert&58.53/24.1 & 43.5/16.07& 35.07/14.57 & 26.7/9.2&- &- &- &-\\
          &W2v-Bert-LM &52.93/23.77 &39.13/19 &33.77/15.7 &30.4/17.2 &- &- &- &-\\
    \hline
    Xhosa &XLS-R  &54.70/10 &19.45/4 &11.42/2 &10.20/2 &8.53/1 & -& -&-\\
          &XLS-R-LM &48.73/10 &18.27/4 &10.43/2 &9.51/2 &7.78/2 &- &- &-\\
          &W2v-Bert  &27.83/5 &17.14/3 &11.04/2 &6.49/1 &7.13/1 &- & -&-\\
          &W2v-Bert-LM &25.32/5 &15.70/3 &10.28/2 &6.19/1 &6.67/1 &- &- &-\\
    \hline
    Zulu &XLS-R  &45.31/8 &24.67/4 &22.51/4 &12.84/2 &9.69/2 & -&- &-\\
         &XLS-R-LM &34.64/7 &20.70/4 &17.93/3 &11.76/2 &8.59/1 &- &- &-\\
         &W2v-Bert  &28.02/5 &19.58/3 &20.38/3 &16.51/3 &7.89/1& -& -&-\\
         &W2v-Bert-LM &24.58/4 &17.95/3 &17.83/3 &15.26/3 &7.40/1 &- &- &-\\
    \hline
  \end{tabular}
\end{table*}

\section{Discussion}\label{sec:discussion}

The comparative evaluation of XLS-R, Whisper, W2v-BERT, and MMS provides several insights into how model architectures, pre-training coverage, and dataset characteristics interact in African low-resource ASR. Although all models improve with additional fine-tuning data, the rate and limits of improvement vary by language and model family, indicating that performance cannot be explained by the size of the dataset alone.

Whisper and MMS demonstrate strong performance in very low-resource settings (1–10 hours). Their large and diverse pre-training corpora likely provide strong linguistic priors that allow them to adapt quickly with minimal fine-tuning. In the case of Whisper, this advantage is partly explained by its encoder–decoder architecture, where the autoregressive decoder already functions as an internal language model. These built-in capabilities support rapid adaptation, but also help explain why Whisper’s gains plateau earlier than those of XLS-R and W2v-BERT, which continue to improve more gradually with larger fine-tuning sets.

Language-specific results show that performance depends on both pre-training representation and dataset characteristics. Languages represented by read-speech corpora, such as Afrikaans, Xhosa, and Zulu from the NCHLT corpus, achieve relatively low WERs across models and benefit from faster convergence. In contrast, spontaneous or conversational corpora present greater challenges. Akan, drawn from the Ashesi finance dataset, retains higher error rates even at larger training sizes, while Bemba and Ewe also struggle due to conversational style, disfluencies, and limited representation in pre-training. These findings suggest a generalizable pattern; read speech corpora enable faster convergence with fewer hours of data, whereas conversational or noisy speech requires more data to stabilize performance and remains harder to model.

The integration of external n-gram language models produced the largest improvements in mid-resource conditions (10–50 hours). For XLS-R and W2v-BERT, notable reductions in WER were observed in several languages, with Luganda, Swahili and Afrikaans showing clear gains when decoding with an LM. However, the benefits were not uniform. Gains diminished once the acoustic models matured beyond 100 hours, and in some cases the performance worsened. For example, Bambara, Shona, and Bemba exhibited instability or regression when LM decoding was applied, likely due to mismatches between the LM training text and the spontaneous or conversational style of the audio. These results indicate that LM decoding is most valuable when the acoustic model is moderately trained and when the text corpus is sufficiently aligned with the speech domain. When either condition is not met, the LM can introduce bias or errors instead of correcting them.

Together, the results highlight three main insights. First, model architecture and pre-training coverage strongly shape data efficiency, with encoder–decoder models favoring rapid adaptation and encoder-only models scaling more gradually with larger fine-tuning sets. Second, external language model decoding is most effective in mid-resource conditions, but its usefulness depends on language and domain alignment. Third, corpus characteristics such as domain and style fundamentally affect outcomes, with read speech datasets enabling faster convergence, and conversational datasets presenting persistent challenges. These findings underscore the importance of selecting models with awareness of their internal capacities, combining acoustic and text resources appropriately, and taking into account domain-specific factors when building ASR systems for low-resource African languages.

\section{Conclusion}
\label{sec:conclusion}

This study presented a comparative evaluation of four pre-trained ASR models, Whisper, MMS, XLS-R, and W2v-BERT, across 13 African languages. By systematically varying the amount of fine-tuning data, we analyzed how performance scales with resource availability and examined the contribution of external n-gram language models. The results show that all models improve with additional data, but their scaling behavior differs: Whisper and MMS adapt quickly under very low resource conditions, while XLS-R and W2v-BERT achieve stronger gains as larger datasets become available. External language models provided consistent benefits in the mid-resource range, though their impact was uneven across languages and in some cases counterproductive, reflecting the importance of domain alignment between text and speech corpora. The findings also highlight broader lessons for low-resource ASR. Pre-training coverage strongly shapes early adaptation, external LM decoding is most effective when acoustic models are moderately trained, and corpus characteristics matter: read speech enables faster convergence, while conversational or spontaneous data require more transcriptions to stabilize performance. Whisper’s encoder-decoder design may also account for its stronger low-resource behavior, suggesting that it implicitly incorporates LM-like capabilities. Beyond this study's analysis, we are extending this work with detailed error analyzes and human evaluations to assess orthographic fidelity, meaning preservation, and error typologies such as substitutions, deletions, insertions, and diacritic handling. In general, this study provides insights that inform model and data choices for African languages while also offering guidance that generalizes to the development of ASR systems in other low-resource settings.

\acks{ Makerere University is grateful for support from the Gates Foundation and Clear Global, who supported this work. We are also grateful to Howard Lakougna at the Gates Foundation and Christian Resch from Clear Global, who provided valuable feedback and partnership during the project. Lastly, we are very grateful to all organisations that provided us with the speech datasets that we used in this paper.  }

\bibliography{pmlr-sample}






\end{document}